# From Influence Diagrams to Junction Trees

**Frank Jensen**     **Finn V. Jensen**     **Søren L. Dittmer**
Department of Mathematics and Computer Science
Aalborg University
Fredrik Bajers Vej 7E, DK-9220 Aalborg Øst, Denmark
E-mail: fj@iesd.auc.dk, fvj@iesd.auc.dk, dittmer@iesd.auc.dk

## Abstract

We present an approach to the solution of decision problems formulated as influence diagrams. This approach involves a special triangulation of the underlying graph, the construction of a junction tree with special properties, and a message passing algorithm operating on the junction tree for computation of expected utilities and optimal decision policies.

## 1 INTRODUCTION

Influence diagrams were introduced by Howard and Matheson (1981) as a formalism to model decision problems with uncertainty for a single decision maker.

The original way to evaluate such problems involved unfolding the influence diagram into a decision tree and using the "average-out and fold-back" algorithm on that tree. Shachter (1986) describes a way to evaluate an influence diagram without tranforming it into a decision tree. The method operates directly on the influence diagram by means of the node-removal and arc-reversal operations. These operations successively transform the diagram, ending with a diagram with only one utility node that holds the utility of the optimal decision policy; the policies for the individual decisions are computed during the operation of the algorithm (when decision nodes are removed).

Shenoy (1992) describes another approach to the evaluation of influence diagrams: the influence diagram is converted to a valuation network, and the nodes are removed from this network by fusing the valuations bearing on the node (variable) to be removed. Shenoys algorithm is slightly more efficient than Shachters algorithm in that it maintains a system of valuations, whereas Shachters algorithm maintains a system of conditional probability functions (in addition to the utility functions), and some extra work (some division operations) is required to keep the probability potentials normalized. Ndilikilikesha (Shachter and Ndilikilikesha, 1993; Ndilikilikesha, 1994) modified the node-removal/arc-reversal algorithm to avoid these extra divisions; the result is an algorithm that is equivalent to Shenoys algorithm with respect to computational efficiency.

Our work builds primarily on the work of Shenoy (1992) and Shachter and Peot (1992), in addition to our previous work on propagation algorithms for the expert system shell Hugin (Andersen et al., 1989; Jensen et al., 1990).

## 2 INFLUENCE DIAGRAMS

An *influence diagram* is a belief network augmented with decision variables and a utility function.

The structure of a decision problem is determined by an acyclic directed graph G. The vertices of G represent either random variables (also known as chance or probabilistic variables) or decision variables, and the edges represent probabilistic dependencies between variables. Decision variables represent actions that are under the full control of the decision maker; hence, we do not allow decision variables to have parents in the graph.

Let $U_R$ be the set of random variables, and let the set of decision variables be $U_D = \{D_1, \ldots, D_n\}$, with the decisions to be made in the order of their index. Let the universe of all variables be denoted by $U = U_R \cup U_D$. We partition $U_R$ into a collection of disjoint sets $I_0, \ldots, I_n$; for $0 < k < n$, $I_k$ is the set of variables that will be observed[1] between decision $D_k$ and $D_{k+1}$; $I_0$ is the initial evidence variables, and $I_n$ is the set of variables that will never be observed (or will be observed after the last decision). This induces a partial order $\prec$ on $U$:

$$I_0 \prec D_1 \prec I_1 \prec \cdots \prec D_n \prec I_n.$$

---
[1] By 'observed,' we mean that the true state of the variable will be revealed.

We associate with each random variable A a conditional probability function $\phi_A = P(A|\mathcal{P}_A)$, where $\mathcal{P}_A$ denotes the set of parents of A in G.

The *state space* $\mathcal{X}_V$ for $V \subseteq U$ is defined as the Cartesian product of the sets of possible outcomes/decision alternatives for the individual variables in V. A *potential* $\phi_V$ for a set V of variables is a function from $\mathcal{X}_V$ to the set of real numbers.

The potential $\phi_V$ can be *extended* to a potential $\phi_W$ ($V \subseteq W$) by simply ignoring the extra variables: $\phi_W(w) = \phi_V(v)$ if $v$ is the projection of $w$ on V.

Given two potentials, $\phi$ and $\psi$. The product $\phi * \psi$ and the quotient $\phi/\psi$ are defined in the natural way, except that $0/0$ is defined to be 0 ($x/0$ for $x \neq 0$ is undefined).

The (a priori) joint probability function $\phi_U$ is defined as

$$\phi_U = \prod_{A \in U_R} \phi_A.$$

For each instance of $U_D$ (i.e., each element of $\mathcal{X}_{U_D}$), $\phi_U$ defines a joint probability function on $U_R$.

A solution to the decision problem consists of a series of decisions that maximizes some objective function. Such a function is called a *utility* function. Without loss of generality, we may assume that the utility function $\psi$ is a potential that may be written as a sum of (possibly) simpler potentials:

$$\psi = \sum_{k=1}^{m} \psi_k$$

We need to impose a restriction on the decision problem, namely that a decision cannot have an impact on a variable already observed. This translates into the property

$$P(I_k|I_0,\ldots,I_{k-1},D_1,\ldots,D_n) = P(I_k|I_0,\ldots,I_{k-1},D_1,\ldots,D_k). \quad (1)$$

In words: we can calculate the joint distribution for $I_k$ without knowledge of the states of $D_{k+1}, \ldots, D_n$ (i.e., the future decisions).

## 2.1 GRAPHICAL REPRESENTATION

In Figure 1, an example of an influence diagram is shown. Random variables are depicted as circles, and decision variables are depicted as squares. Moreover, each term of the utility function is depicted as a diamond, and the domain of the term is indicated by its parent set. The partial order $\prec$ is indicated by making $I_{k-1}$ the parent set of $D_k$, and we shall use the convention that the temporal order of the decisions are read from left to right.

The independence restriction imposed on the decision problem can be verified by checking that, in the influence diagram, there is no directed path from a decision $D_k$ to a decision $D_i$ ($i < k$).

## 3 DECISION MAKING

Assume we have to choose an alternative for decision $D_n$ (i.e., the last decision). We have already observed the random variables $I_0, \ldots, I_{n-1}$, and we have chosen alternatives for decisions $D_1, \ldots, D_{n-1}$. The *maximum expected utility principle*[2] says that we should choose the alternative that maximizes the expected utility. The maximum expected utility for decision $D_n$ is given by

$$\rho_n = \max_{D_n} \sum_{I_n} P(I_n|I_0,\ldots,I_{n-1},D_1,\ldots,D_n) * \psi.$$

Obviously, $\rho_n$ is a function of previous observations and decisions. We calculate the maximum expected utility for decision $D_k$ ($k < n$) in a similar way:

$$\rho_k = \max_{D_k} \sum_{I_k} P(I_k|I_0,\ldots,I_{k-1},D_1,\ldots,D_k) * \rho_{k+1}. \quad (2)$$

We note that $\rho_k$ is well-defined because of (1).

By expansion of (2), we get

$$\rho_k = \max_{D_k} \sum_{I_k} P(I_k|I_0,\ldots,I_{k-1},D_1,\ldots,D_k)$$
$$* \max_{D_{k+1}} \sum_{I_{k+1}} P(I_{k+1}|I_0,\ldots,I_k, D_1,\ldots,D_{k+1}) * \rho_{k+2}$$
$$= \max_{D_k} \sum_{I_k} \max_{D_{k+1}} \sum_{I_{k+1}} P(I_k, I_{k+1}|I_0,\ldots,I_{k-1}, D_1,\ldots,D_{k+1}) * \rho_{k+2}.$$

The last step follows from (1) and the chain rule of probability theory: $P(A|B,C)P(B|C) = P(A,B|C)$. By further expansion, we get

$$\rho_k = \max_{D_k} \sum_{I_k} \cdots \max_{D_n} \sum_{I_n} P(I_k,\ldots,I_n|I_0,\ldots,I_{k-1}, D_1,\ldots,D_n) * \psi.$$

From this formula, we see that in order to calculate the maximum expected utility for a decision, we have to perform a series of marginalizations (alternately sum- and max-marginalizations), thereby eliminating the variables.

When we eliminate a variable A from a function $\phi$, expressible as a product of simpler functions, we partition the factors into two groups: the factors that involve A, and the factors that do not; call (the product of) these factors $\phi_A^+$ and $\phi_A^-$, respectively. The marginal $\sum_A \phi$ is then equal to $\phi_A^- * \sum_A \phi_A^+$; $\sum_A \phi_A^+$

---
[2] There are good arguments for adhering to this principle. See, e.g., (Pearl, 1988).

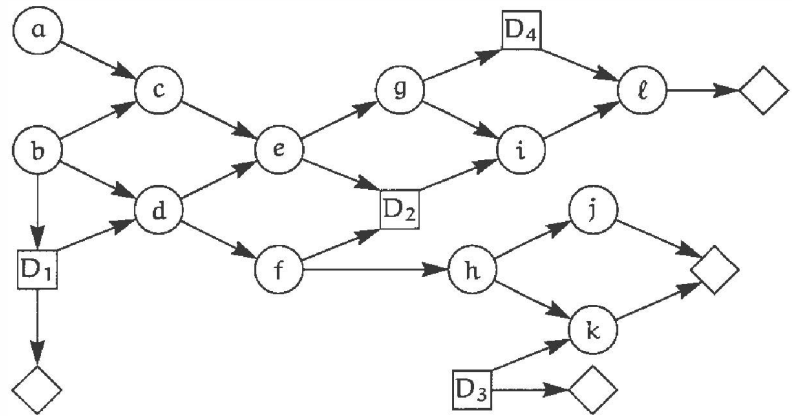

FIGURE 1.
An influence diagram for a decision problem with four decisions. The set of variables is partitioned into the sets: $I_0 = \{b\}$, $I_1 = \{e, f\}$, $I_2 = \emptyset$, $I_3 = \{g\}$, and $I_4 = \{a, c, d, h, i, j, k, \ell\}$. The utility function is a sum of four local utilities, three of which are associated with single variables ($D_1$, $D_3$, and $\ell$), and one associated with the pair $(j, k)$.

then becomes a new factor that replaces the product $\phi_A^+$ in the expression for $\phi$. This also holds true for max-marginalizations, provided $\phi_A^-$ does not assume negative values.

The product $\phi$ may be represented by an undirected graph, where each maximal complete set (clique) of nodes (the nodes being variables) corresponds to a factor (or a group of factors) of $\phi$ with that set as its domain. Marginalizing a variable A out of $\phi$ then corresponds to the following operation on the graph: the set of neighbors of A in the graph is completed, and A is removed. It is a well-known result that all variables can be eliminated in this manner without adding edges if and only if the graph is triangulated (Rose, 1970).

Obviously, it is desirable to eliminate all variables without adding extra edges to the graph since this means that we do not create new factors with a larger domain than the original factors (the complexity of representing and manipulating a factor is exponential in the number of variables comprising its domain). However, in most cases, this is not possible: we have to add some edges, and the elimination order chosen will determine how many and hence also the size of the cliques. Unfortunately, it is $\mathcal{NP}$-hard to find an optimal elimination order for all reasonable criteria of optimality.

When we perform inference in a belief network (i.e., calculation of the marginal probability of some variable given evidence on other variables), the computation only involves sum-marginalizations. In this case, we can eliminate the variables in any order, since the order of two marginalizations of the same kind can be interchanged. However, the calculation of the maximum expected utility involves both max- and sum-marginalizations; but — in general — we cannot interchange the order of a max- and a sum-marginalization; this fact imposes some restrictions on the elimination order.

## 4 COMPILATION OF INFLUENCE DIAGRAMS

We first form the moral graph of G. This means adding (undirected) edges between vertices with a common child. We also complete the vertex sets corresponding to the domains of the utility potentials. Finally, we drop directions on all edges.

Next, we triangulate the moral graph in such a way that it facilitates the computation of the maximum expected utility. This is equivalent to the selection of a special elimination order for the moral graph: the reverse of the elimination order must be some extension of $\prec$ to a total order.

Finally, we organize the cliques of the triangulated graph in a *strong junction tree*: A tree of cliques is called a *junction tree* if for each pair $(C_1, C_2)$ of cliques, $C_1 \cap C_2$ is contained in every clique on the path connecting $C_1$ and $C_2$. For two adjacent cliques, $C_1$ and $C_2$, the intersection $C_1 \cap C_2$ is called a *separator*. A junction tree is said to be *strong* if it has at least one distinguished clique R, called a *strong root*, such that for each pair $(C_1, C_2)$ of adjacent cliques in the tree, with $C_1$ closer to R than $C_2$, there exists an ordering of $C_2$ that respects $\prec$ and with the vertices of the separator $C_1 \cap C_2$ preceding the vertices of $C_2 \setminus C_1$. This property ensures that the computation of the maximum expected utility can be done by local message passing in the junction tree (see Section 5).

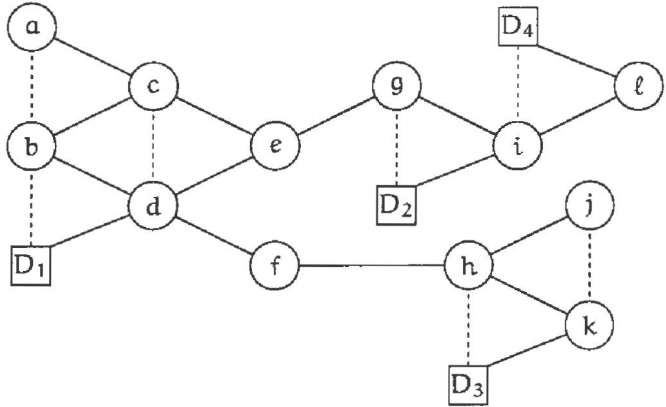

**FIGURE 2.**
The moral graph for the decision problem in Figure 1. Edges added by the moralization process are indicated by dashed lines.

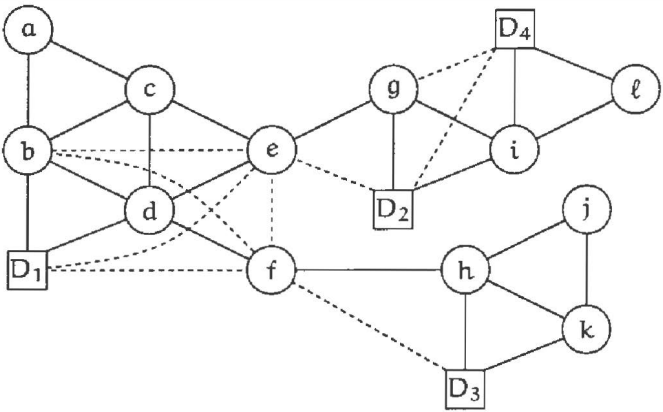

**FIGURE 3.**
The triangulated graph of the moral graph in Figure 2. Fill-in edges added during triangulation are indicated by dashed lines.

### 4.1 CONSTRUCTION OF STRONG JUNCTION TREES

Let $\alpha$ be a numbering of $U$ (i.e., a bijection $\alpha: U \leftrightarrow \{1,\ldots,|U|\}$) such that for all $u, v \in U$, $u \prec v$ implies $\alpha(u) < \alpha(v)$. We assume that $\alpha$ is the elimination order used to produce the triangulated graph of $G$: vertices with higher numbers are eliminated before vertices with lower numbers.

Let $C$ be a clique of the triangulated graph, and let $v \in C$ be the highest-numbered vertex such that the vertices $\{w \in C \mid \alpha(w) < \alpha(v)\}$ have a common neighbor $u \notin C$ with $\alpha(u) < \alpha(v)$. If such a vertex $v$ exists, we define the *index* for $C$ as $\text{index}(C) = \alpha(v)$; otherwise, we define $\text{index}(C) = 1$. Intuitively, the index for a clique $C$ identifies the step in the elimination process that causes $C$ to "disappear" from the graph. It is easy to see that the index for a clique $C$ is well-defined, and that no two cliques have the same index. Moreover, unless $\text{index}(C) = 1$, the set $\{v \in C \mid \alpha(v) < \text{index}(C)\}$ will be a proper subset of some other clique with a lower index than $C$.

Let the collection of cliques of the triangulated graph be $C_1, \ldots, C_m$, ordered in increasing order according to their index. As a consequence of the above construction, this ordering will have the *running intersection property* (Beeri et al., 1983), meaning that

$$\text{for all } k > 1: \ S_k = C_k \cap \bigcup_{i=1}^{k-1} C_i \subseteq C_j \ \text{ for some } j < k.$$

It is now easy to construct a strong junction tree: we start with $C_1$ (the root); then we successively attach each clique $C_k$ to some clique $C_j$ that contains $S_k$.

Consider the decision problem in Figure 1. Figure 2 shows the moral graph for this problem: edges have been added between vertices with a common child (including utility vertices), utility vertices have been removed, and directions on all edges have been dropped. Note that the time precedence edges leading into decision vertices are not part of the graph and are thus not shown.

Figure 3 shows the strong triangulation of the graph in Figure 2 generated by the elimination sequence $\ell$, $j$, $k$, $i$ (fill-ins: $D_2 \sim D_4$ and $g \sim D_4$), $h$ (fill-in: $f \sim D_3$), $a$, $c$ (fill-in: $b \sim e$), $d$ (fill-ins: $D_1 \sim e$, $D_1 \sim f$, $b \sim f$, and $e \sim f$), $D_4$, $g$ (fill-in: $e \sim D_2$), $D_3$, $D_2$, $f$, $e$, $D_1$, and $b$. This graph has the following cliques: $C_{16} = \{D_4, i, \ell\}$, $C_{15} = \{h, k, j\}$, $C_{14} = \{D_3, h, k\}$, $C_{11} = \{b, c, a\}$, $C_{10} = \{b, e, d, c\}$, $C_8 = \{D_2, g, D_4, i\}$, $C_6 = \{f, D_3, h\}$, $C_5 = \{e, D_2, g\}$, and $C_1 = \{b, D_1, e, f, d\}$. Using the above algorithm, we get the strong junction tree shown in Figure 4 for this collection of cliques. (There exists another strong junction tree for this collection, obtained by replacing the edge $C_5 \to C_1$ by the edge $C_5 \to C_{10}$. This tree is computationally slightly more efficient, but — unfortunately — it cannot be constructed by the algorithm given in this paper.)

In general, previous observations and decisions will be relevant when making a decision. However, sometimes only a subset of these observations and decisions are needed to make an optimal decision. For example, for the decision problem in Figure 1, the variable $e$ summarizes all relevant information available when decision $D_2$ has to be made: although $f$ is observed just before decision $D_2$, it has no relevance for that decision (it does, however, have relevance for decision $D_3$). This fact is detected by the compilation algorithm: the only link from $D_2$ to past observations and decisions goes to $e$.

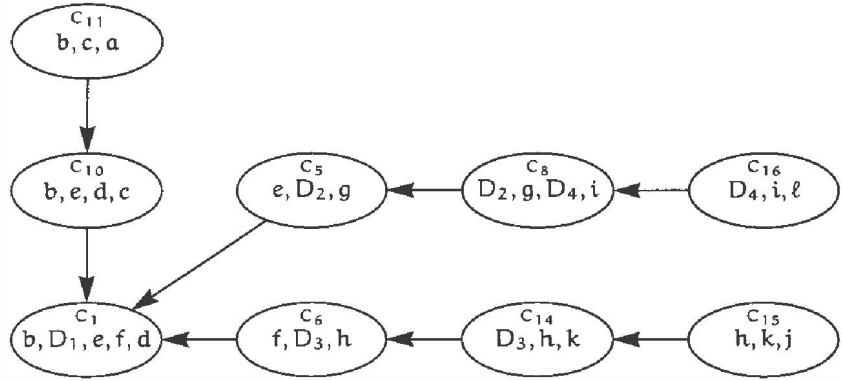

FIGURE 4. A strong junction tree for the cliques of the graph in Figure 3.

## 5  USING THE STRONG JUNCTION TREE FOR COMPUTATIONS

We perform computations in the junction tree as a special 'collect' operation from the leaves of the junction tree to some strong root of the tree.

To each clique C in the junction tree, we associate a probability potential $\phi_C$ and a utility potential $\psi_C$ defined on $\mathcal{X}_C$. Let $\mathcal{C}$ be the set of cliques. We define the joint potentials $\phi$ and $\psi$ for the junction tree as

$$\phi = \prod_{C \in \mathcal{C}} \phi_C; \qquad \psi = \sum_{C \in \mathcal{C}} \psi_C.$$

We initialize the junction tree as follows: each variable $A \in U_R$ is assigned to a clique that contains $A \cup \mathcal{P}_A$. The probability potential for a clique is the product of the conditional probability functions for the variables assigned to it. For cliques with no variables assigned to them, the probability potentials are unit functions. In this way, the joint probability potential for the junction tree becomes equal to the joint probability function for the influence diagram. Similarly, each utility function $\psi_k$ is assigned to some clique that can accommodate it. The utility potential for a clique is the sum of the utility functions assigned to it; for cliques with no utility functions assigned to them, the utility potential is a null function.

We need a generalized marginalization operation that acts differently on random and decision variables. We denote the operation by '$\mathsf{M}$'. For random variable A and decision variable D, we define

$$\mathsf{M}_A \phi = \sum_A \phi; \qquad \mathsf{M}_D \phi = \max_D \phi.$$

For a set V of variables, we define $\mathsf{M}_V \phi$ as a series of single-variable marginalizations, in the inverse order as determined by the relation $\prec$. Note that although $\prec$ is only a partial order, $\mathsf{M}_V \phi$ is well-defined.

Now, let T be a strong junction tree, and let $C_1$ and $C_2$ be adjacent cliques with separator S in T. We say that $C_1$ *absorbs* from $C_2$ if $\phi_{C_1}$ and $\psi_{C_1}$ change to $\phi'_{C_1}$ and $\psi'_{C_1}$ as follows:

$$\phi'_{C_1} = \phi_{C_1} * \phi_S; \qquad \psi'_{C_1} = \psi_{C_1} + \frac{\psi_S}{\phi_S};$$

where

$$\phi_S = \mathsf{M}_{C_2 \setminus S} \phi_{C_2}; \qquad \psi_S = \mathsf{M}_{C_2 \setminus S} \phi_{C_2} * \psi_{C_2}.$$

Note that this definition of absorption is 'asymmetric' in the sense that information only flows in the direction permitted by the partial order $\prec$. It is possible to generalize this definition of absorption to a symmetric definition similar to the one given in (Jensen et al., 1990) for the case of pure probabilistic influence diagrams.

Clearly, the complexity of an absorption operation is $O(|\mathcal{X}_{C_1}| + |\mathcal{X}_S| + |\mathcal{X}_{C_2}|)$. Note in particular that the contribution from the division operation plays a much smaller rôle than in (Shenoy, 1992), since division operations are performed on separators only.

We will need the following lemma, which we shall state without proof.

**Lemma 1** *Let D be a decision variable, and let V be a set of variables that includes all descendants of D in G. Then $\mathsf{M}_{V \setminus \{D\}} \phi u$, considered as a function of D alone, is a non-negative constant.*

Let T be a strong junction tree with at least two cliques; let $\phi_T$ be the joint probability potential and $\psi_T$ the joint utility potential on T. Choose a strong root R for T and some leaf L ($\neq$ R); let T \ L denote the strong junction tree obtained by absorbing L into its neighbor N and removing L; denote the separator between N and L by S.

**Theorem 1** *After absorption of L into T, we have*

$$\mathop{\mathrm{M}}_{L\setminus S} \phi_T * \psi_T = \phi_{T\setminus L} * \psi_{T\setminus L}.$$

*Proof:* Let

$$\overline{\phi}_L = \prod_{C\in\mathcal{C}\setminus\{L\}} \phi_C; \quad \overline{\psi}_L = \sum_{C\in\mathcal{C}\setminus\{L\}} \psi_C.$$

Since $\overline{\phi}_L$ does not assume negative values, we get

$$\mathop{\mathrm{M}}_{L\setminus S} \phi_T * \psi_T = \overline{\phi}_L * \mathop{\mathrm{M}}_{L\setminus S} \phi_L * (\psi_L + \overline{\psi}_L).$$

We have to show that

$$\mathop{\mathrm{M}}_{L\setminus S} \phi_L * (\psi_L + \overline{\psi}_L) = \phi_S * \left(\frac{\psi_S}{\phi_S} + \overline{\psi}_L\right),$$

where

$$\phi_S = \mathop{\mathrm{M}}_{L\setminus S} \phi_L; \quad \psi_S = \mathop{\mathrm{M}}_{L\setminus S} \phi_L * \psi_L.$$

We shall prove this by induction. Let $X_1, \ldots, X_\ell$ be some ordering of $L \setminus S$ that respects $\prec$. Now, consider the equation:

$$\mathop{\mathrm{M}}_{X_k} \cdots \mathop{\mathrm{M}}_{X_\ell} \phi_L * (\psi_L + \overline{\psi}_L) = \phi^{(k)} * \left(\frac{\psi^{(k)}}{\phi^{(k)}} + \overline{\psi}_L\right), \quad (3)$$

where

$$\phi^{(k)} = \mathop{\mathrm{M}}_{X_k} \cdots \mathop{\mathrm{M}}_{X_\ell} \phi_L; \quad \psi^{(k)} = \mathop{\mathrm{M}}_{X_k} \cdots \mathop{\mathrm{M}}_{X_\ell} \phi_L * \psi_L.$$

(For $k = 1$, (3) is equivalent to the desired result.) For $k > \ell$, (3) is clearly true; for $1 \leq k \leq \ell$, we have two cases:

(1) $X_k$ is a random variable. By induction, we get

$$\mathop{\mathrm{M}}_{X_k} \cdots \mathop{\mathrm{M}}_{X_\ell} \phi_L * (\psi_L + \overline{\psi}_L)$$
$$= \sum_{X_k} \phi^{(k+1)} * \left(\frac{\psi^{(k+1)}}{\phi^{(k+1)}} + \overline{\psi}_L\right)$$
$$= \sum_{X_k} \psi^{(k+1)} + \overline{\psi}_L * \sum_{X_k} \phi^{(k+1)}$$
$$= \psi^{(k)} + \overline{\psi}_L * \phi^{(k)} = \phi^{(k)} * \left(\frac{\psi^{(k)}}{\phi^{(k)}} + \overline{\psi}_L\right)$$

The correctness of the last step follows from the fact that $\phi^{(k)}(x) = 0$ implies $\psi^{(k)}(x) = 0$ (so that our division-by-zero convention applies).

(2) $X_k$ is a decision variable. By induction, we get

$$\mathop{\mathrm{M}}_{X_k} \cdots \mathop{\mathrm{M}}_{X_\ell} \phi_L * (\psi_L + \overline{\psi}_L)$$
$$= \max_{X_k} \phi^{(k+1)} * \left(\frac{\psi^{(k+1)}}{\phi^{(k+1)}} + \overline{\psi}_L\right).$$

Because of Lemma 1, $\phi^{(k+1)}$, considered as a function of $X_k$ alone, is a non-negative constant, and we get

$$\left(\max_{X_k} \phi^{(k+1)}\right) * \left(\frac{\max_{X_k} \psi^{(k+1)}}{\max_{X_k} \phi^{(k+1)}} + \overline{\psi}_L\right)$$
$$= \phi^{(k)} * \left(\frac{\psi^{(k)}}{\phi^{(k)}} + \overline{\psi}_L\right). \blacksquare$$

By successively absorbing leaves into a strong junction tree, we obtain probability and utility potentials on the intermediate strong junction trees that are equal to the marginals of the original potentials with respect to the universes of these intermediate trees. This is ensured by the construction of the junction tree in which variables to be marginalized out early are located farther away from the root than variables to be marginalized out later.

The optimal policy for a decision variable can be determined from the potentials on the clique that is closest to the strong root and contains the decision variable (that clique may be the root itself), since all variables that the decision variable may depend on will also be members of that clique.

For our example decision problem (Figure 4), we can determine the optimal policy for $D_1$ from (the potentials on) clique $C_1$ (the root), and the optimal policies for the remaining decisions can be determined from cliques $C_5$ (decision $D_2$), $C_6$ (decision $D_3$), and $C_8$ (decision $D_4$).

If only the maximum expected utility is desired, it should be noted that only storage for the 'active' part of the junction tree during the collect operation needs to be reserved; this means that storage for at most two adjacent cliques and each clique that corresponds to a branch point on the currently active path from the root to a leaf must be reserved. Since elimination of a group of variables can be implemented more efficiently than the corresponding series of single-variable eliminations, it is still useful to organize the computations according to the structure of the strong junction tree as compared to (Shenoy, 1992).

## 6  CONCLUSION

We have described an algorithm to transform a decision problem formulated as an influence diagram into a secondary structure, a strong junction tree, that is particularly well-suited for efficient computation of maximum expected utilities and optimal decision policies. The algorithm is a refinement of the work by Shenoy (1992) and Shachter and Peot (1992); in particular, the construction of the strong junction tree and its use for computations has been elaborated upon.

The present work forms the basis for an efficient computer implementation of Bayesian decision analysis in the expert system shell Hugin (Andersen et al., 1989).

We have not given an algorithm to construct the elimination sequence that generates the strong triangulation. However, the triangulation problem is simpler than for ordinary probability propagation, since the set of admissible elimination sequences is smaller; at this stage, it appears that simple adaptations of the heuristic algorithms described by Kjærulff (1990) work very well. Moreover, even given a triangulation, there might exist several strong junction trees for the collection of cliques.

Besides the use of the strong junction tree for computation of expected utilities and optimal decision policies, it should be possible to exploit the junction tree for computation of probabilities for random variables that only depend on decisions that have already been made. Ideally, this should be done through a 'distribute' operation from the root towards the leaves of the junction tree.

Work regarding these problems is in progress.

### Acknowledgements

This work has been partially funded by the Danish research councils through the PIFT programme.